\documentclass[journal]{IEEEtran}

\ifCLASSINFOpdf
\else
\fi
\usepackage{array}
\usepackage{times,amsmath,graphicx,comment,url,color,float,subfig,hyperref,cite}
\usepackage{algorithmic, algorithm}
\usepackage{multirow}
\usepackage{tabularx}

\newcommand{\mycomment}[1]{}

\begin{document}
%
\title{Next Generation Business Intelligence and Analytics: A Survey}
%
%
%
\author{
\IEEEauthorblockN{Quoc Duy Vo\IEEEauthorrefmark{1} Jaya Thomas\IEEEauthorrefmark{1} Shinyoung Cho\IEEEauthorrefmark{1} Pradipta De\IEEEauthorrefmark{2} Bong Jun Choi\IEEEauthorrefmark{1} Lee Sael\IEEEauthorrefmark{1} \\}
\IEEEauthorblockA{\IEEEauthorrefmark{1}Department of Computer Science, SUNY Korea, Incheon, South Korea\\
Department of Computer Science, Stony Brook University, New York, USA\\
Email: {\{rayvo, jaya.thomas, sycho, sael, bjchoi\}@sunykorea.ac.kr}} \\ 
\IEEEauthorblockA{\IEEEauthorrefmark{2}Department of Computer Sciences, Georgia Southern University, Georgia, USA\\
Email:{pde@georgiashouthern.edu}}
}

\markboth{IEEE Communications Surveys \& Tutorials}%
{}

%



\maketitle

\begin{abstract}
Business Intelligence and Analytics (BI\&A) is the process of extracting and predicting business-critical insights from data.
Traditional BI focused on data collection, extraction, and organization to enable efficient query processing for deriving insights from historical data.
With the rise of big data and cloud computing, there are many challenges and opportunities for the BI.
Especially with the growing number of data sources, traditional BI\&A are evolving to provide intelligence at different scales and perspectives - operational BI, situational BI, self-service BI.
In this survey, we review the evolution of business intelligence systems in full scale from back-end architecture to and front-end applications.
We focus on the changes in the back-end architecture that deals with the collection and organization of the data. We also review the changes in the front-end applications, where analytic services and visualization are the core components.
Using a uses case from BI in Healthcare, which is one of the most complex enterprises, we show how BI\&A will play an important role beyond the traditional usage.
The survey provides a holistic view of Business Intelligence and Analytics for anyone interested in getting a complete picture of the different pieces in the emerging next generation BI\&A solutions. 
\end{abstract}


%
\IEEEpeerreviewmaketitle

\section{Introduction}
\label{intro:sec}

Business Intelligence can be expressed as the automated process to collect raw data from heterogeneous sources, and to organize them in a systematic manner.
With the automated processes, models and insights can be derived from the data to improve business processes.
The best practice in enterprise BI architectures is to split back-end architecture, associated with the data collection and data organization, with the the front-end, where data analyzed and displayed to the user.
The transactions data are generated when transactions are processed and they are stored in the Online Transaction Processing server (OLTP), also called Operational Data Sources.
From the OLTP servers, data is extracted, transformed, and stored into a data warehouse, which is a structured data repository.
Different query optimization techniques can be applied on the data warehouse for speed-up of data analysis and analytics query can run on the data warehouse. Further speed-up can be achieved through creation of data marts, which are subsets of the data warehouse.

\begin{figure*}
\centering
\includegraphics[width=0.7\textwidth]{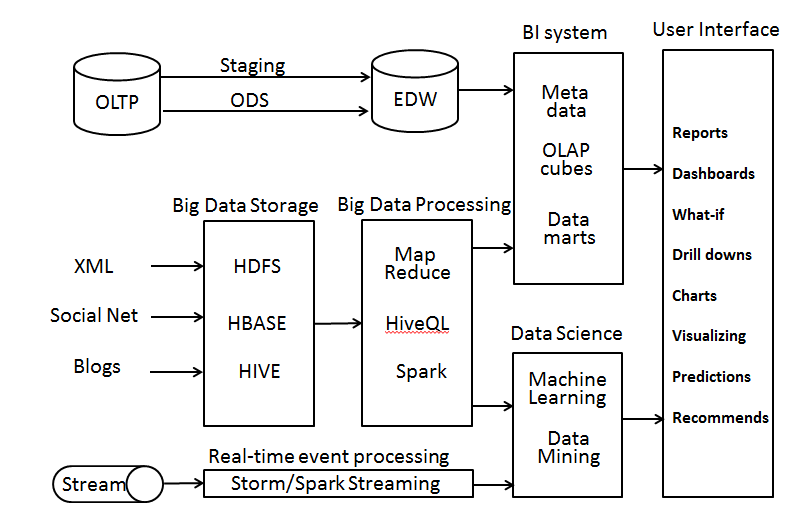}
\caption{Different components in traditional BI architecture.}
\label{components:fig}
\end{figure*}

In addition to the traditional data sources, i.e. transaction data, the sources of BI data are evolving to include even the messages sent via company intranets and personal profiles of employees and customers from the web. The mobile devices and other sensor data also add to the data sources.
However, many of these data sources are not structured.
Texts from messages posted on online social networks and data from different sensors are two type of unstructured data.
This makes it challenging to maintain as traditional relational database while still achieving query efficiency.
In the perspective of the data analysis, more data means more opportunity for the analytics engines to discover more insight. However, there still remains the big data challenges even in the analytic perspective.

The increase in data opens up opportunities in expanding the scope of BI. That is going beyond being just a mechanism to analyze trends from historical data. Situational BI can combine real time data from sensors and other personal information in real time to infer insights that are not traditionally available \cite{2009:situationalBI}. Operational BI is concerned with providing real time insights to the business operations, such as a call center operative who may benefit by getting instant feedback on their work.
Analytics is also evolving with the notion of self-service BI, where the user may compose the analytics rules based on meta-information about the data exposed to her.
These new approaches to BI however, must be carefully orchestrated such that the enterprise governance and compliance models are not violated.

In this survey, we capture the shifting trends in BI architecture. For the backend, we show how different technology evolutions are transforming the architecture. For the frontend, where analytics engines play the pivotal role, we focus on different trends in Machine Learning that are enabling the evolution of BI from the traditional historical analysis tool. We also chart the how challenges in enforcing the enterprise governance models are being addressed in this landscape.

Business Intelligence is no longer just a tool to support enterprise environments. It can be used by public enterprises and by the government to understand social scale initiatives and predict requirements.
We showcase healthcare use case to illustrate how the evolved BI architectures fit into the use cases.
Two other technology trends, namely big data and cloud computing, are also closely tied to the changes happening in the BI architecture.
We present the connections to big data and use of cloud computing as opportunity, and discuss research challenges. 
\section{Preliminaries}
\label{preliminaries:sec}

\subsection{Traditional BI}

\begin{figure}
	\centering
	\includegraphics[width=0.5\textwidth]{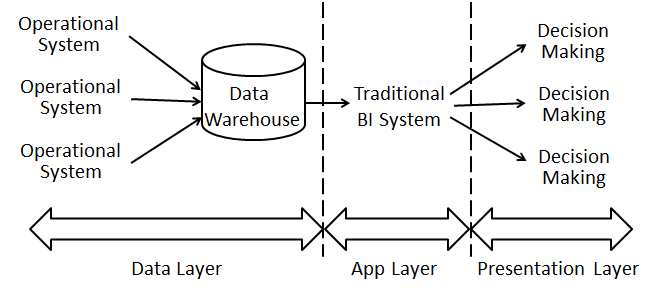}
	\caption{A traditional BI system}
	\label{traditional_bi:fig}
\end{figure}

Traditional BI systems use reporting mechanisms to access transaction data stored in data warehouse. 
Analyzing transaction data can help us to detect patterns and predict business trends.
A traditional BI system consists of three separated layers as shown in figure \ref*{traditional_bi:fig}: presentation layer, application layer, and database layer
With the three-tier architecture, it is challenging to fulfill service level objectives such as maximal response time and minimal throughput rates.
This is due to the difficulties in predicting execution times where the application layer does not know about the data storage management at the low-lever layers.

Although, typical BI system can give us a forward view of the business, it is well-known that traditional BI systems are slow, rigid, time-consuming, and maintenance requires a expert knowledge.
Many researches has been conducted towards improvement of the three-tier architecture as well as to add modern features of the next generation BI.

\subsection{Modern Features of Next Generation BI}
\subsubsection{Operational (Real-Time) BI}
The competitive pressure of today's businesses has increased the need for near real-time BI, also called operational BI.
The goal of operational BI is to reduce the latency between the data acquisition time and data analysis time.
Reduction on the latency enables the system to take appropriate actions when an event occurs.
With the operational BI realized, businesses can detect the patterns or temporal trends over the streaming operational data.

\subsubsection{Situational BI}
Situational BI enable situational awareness. Situation BI is important in companies were fast shift of situations, often external business trends, affect the business \cite{SBI:2009}.
However, such external information, which mostly come from the corporate intranet, external vendor, or internet, are unstructured.
Moreover, these unstructured data need to be integrated with structured information from local data warehouse to support decision marking in real-time.
For example, a business might want to know whether its customers are posting positive or negative comments about its new product.
With the analysis of the comment, businesses can provide immediate feedback to the development team to make the product more competitive.
As another example, it is important for a company to know whether a natural disaster has affected its contracted suppliers. Being aware of the natural disasters, enable business operatives to take appropriate actions necessary in minimizing the loss \cite{Castellanos:2010}.

\subsubsection{Self-service BI}
Self-service BI (SSBI) enables end users to create analytical queries and reports without the IT department's involvement.
The user interface in SSBI applications must be user-friendly, intuitive and easy to use, so that a technical knowledge of the data warehouse is not required.
The user also should be allowed to access or extend not just IT-curated data sources, but also non-traditional ones. 
\subsection{Data Architecture}
\subsubsection{Background and Challenges}
Traditional architecture of business applications consists of three separated layers: presentation, application, and database.
With the three-tier architecture the execution time is hard to predict, due to the correlation between low-level data management operations and high-level processes.
The workload management solutions are usually built on top of general-purpose database management systems, which require time delays when executing requests in parallel.
This creates challenges for modern business applications to be able to work as operational or real-time BI.
Therefore, technologies that enable performing analytical queries and business transaction queries at the same time on the same data is important.

Today's enterprises use an extraction, transformation, and load (ETL) model to extract data, perform transformations, and load the transformed data into the data warehouse.
This model rely on two types of processes which are vital to business operations: online transaction processing (OLTP) and online analytical processing (OLAP).
OLTP is used to manage business processes, such as order processing. OLAP is used to support strategic decision making, such as sales analytics.

\begin{figure}
	\centering
	\includegraphics[width=0.5\textwidth]{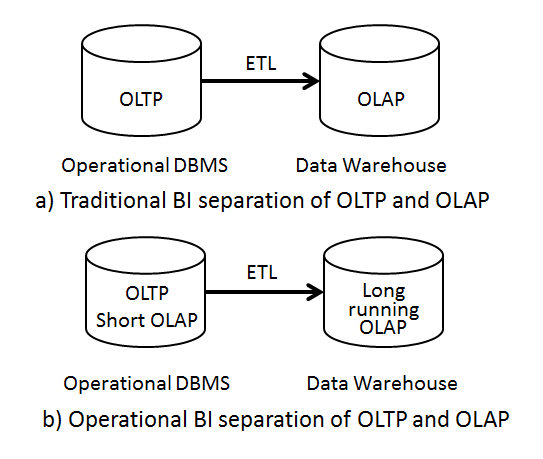}
	\caption{A classification of database systems}
	\label{architecture:fig}
\end{figure}

Workloads from OLTP and OLAP are traditionally executed on the same database system. However OLAP workloads composes mostly of massive read-only operations on the data that is constantly being updated by the OLTP.
Therefore, when both workloads executed in a single database, the transaction processing performance might be unpredictable due to resource contention.
It is thus beneficial to separated out the workloads for OLTP and OLAP.
Figure \ref{architecture:fig}-a shows the traditional ETL-based BI where OLTP and OLAP are separate.
In this architecture, each OLAP workload has to wait until the data in date warehouse are completely updated and visible causing delays. 	

To reduce the delay, today's operational BI systems perform OLTP and shorter-running analytical queries, called short OLAP workloads, together on the operational database management system (DBMS), as shown in Figure \ref{architecture:fig}-b.
However, many short OLTP transactions, which make changes to the database, may conflict with longer-running OLAP workloads.
High synchronization overhead is required to handle the resource contention, which results in low overall resource utilization.

In addition, commercial DBMS use special techniques, such as shadow copy \cite{elmasri2014fundamentals}, to handle mixed workloads with low performance overhead.
That is, different workloads are separated and executed on different logical copies of the data.
This may cause additional space overhead, which increase the infrastructure requirements and costs.
Therefore, managing these mixed workloads (OLTP and OLAP) in the data management systems is a big challenge for current disk-based DBMSs \cite{2010:kuno}.

\subsection{Recent BI systems}
\subsubsection{Extended traditional BI systems}
In this section, we present existing traditional BI techniques which can perform OLTP transactions and OLAP queries on the same database without interfering with each other.
Due to the contradiction of the dramatic explosion of the dynamic data volume, the integration of these mixed workloads on the same system requires extreme performance improvements.

\begin{itemize}
	\item {\em In-memory database (IMDB) }
	
	
	In most today's BI systems, the mixed workload comprised of OLTP and OLAP on a single system can be handled  by using in-memory (or main-memory) database (IMDB).
	This technique requires the system to store all data in main-memory, which is faster than disk-optimized databases since the internal optimization algorithms are simpler and uses fewer CPU instruction.
	When querying the data, this approach provides faster and more predictable performance than disk by eliminating the seek time.
	
	However, IMDB systems can be said to lack of durability due to the losing of stored information when the device loses power or is reset.
	Many IMDB systems have proposed different mechanisms to support durability such as snapshot files, transaction logging, non-volatile DIMM, non-volatile random access memory, and high availability.
	Table \ref{techniques:tb} shows recent BI systems which use different approaches to keep most or all data in main-memory.
	
	\begin{center}
		\begin{table}[hpt]
			{\small
				\hfill{}
				\begin{tabular} {|p{0.8cm}|l|p{2cm}|p{2.6cm}|c|}
					\hline
					System  & Type & Approaches & Achievements  & Year \\ [1ex]
					\hline
					H-store \cite{HStore:2008}& IMDB & Distributed, row-store technique & high OLTP throughput rate & 2008
					\\
					\hline
					Radu Stoica \cite{Stoica:2013}&  Hybrid & Data re-organization,  & High performance & 2013 \\
					&   &  & Reduce paging I/O, and improve memory hit rate & 	\\
					\hline
					Siberia \cite{Siberia:2014}&  Hybrid & Cold data access and migration mechanisms  & Acceptable access rates with 7-14\% throughput loss & 2014
					\\
					\hline
					
				\end{tabular}}
				\hfill{}
				\caption{A summary of traditional BI systems which handle mixed workloads (OLTP and OLAP).}
				\label{techniques:tb}
			\end{table}
		\end{center}
		

		As shown in the table, the BI systems use different approaches to keep most or all data in main-memory to obtain high OLTP throughput rates.
		For example, the H-Store system operates on a distributed cluster of shared-nothing machines where the data resided entirely in main memory.
		By removing traditional DBMS features, such as buffer management, locking and latching, the H-Store system can execute transaction processing with high throughput rates.
		The H-Store prototype was recently commercialized by a start-up company named VoltDB \cite{voltdb}.
		
		\item {\em Hybrids with on-disk database}
		
		Although main-memory is becoming large enough to handle most OLTP database, it may not always be the best option.
		Using the access patterns of the OLTP workloads, where some records are "hot" (frequently accessed), others are "cold" (infrequently or never accessed),
recent systems tend to store the coldest records on a fast secondary storage devices, and hot records should reside in memory to guarantee good performance.
		For example, Stoica and Ailamaki \cite{Stoica:2013} proposed method to migrate data of main-memory database to a larger and cheaper secondary storage.
		In their work, in order to reduce OS paging I/O and improve the main memory hit rates, the relational data structures are re-organized using the access statistics of the OLTP workloads.
		
		
		Recently, Siberia has been introduced as a framework for managing cold data in the Microsoft Hekaton IMDB engine \cite{Siberia:2014}.
		Similar to \cite{Stoica:2013}, it does not require a database be stored entirely in main memory.
		Only some tables can be declared in the main-memory and managed by Hekaton.
		Hekaton focus on how to migrate records to and from the cold store and how to access and update records in the cold store in a transactionally consistent manner.
		The experiment evaluation shows that when the cold store resides on a commodity flash, the Siberia could lead to an acceptable throughput loss of 7-14\%, given that the cold data access rates appropriate for a main-memory optimized database.
	\end{itemize}
	
	\subsubsection{BI Systems with modern features}
	In this section, we describe three modern BIs: operational BI, situational BI, and self-service BI.
	Table. \ref{bi_features:tb} shows a categorization of recent BI systems in terms of modern features.
	\begin{center}
		\begin{table*}[hpt]
			{\small
				\hfill{}
				\begin{tabular} {|p{1.2cm}|p{4cm}|p{4cm}|p{1cm}|p{1cm}|p{1cm}|c|}
					\hline
					System  & Approaches & Achievements & Opera-tional BI & Situa-tional BI  & Self-Service BI & Year \\ [1ex]
					\hline
					HyPer \cite{Kemper:2011}&  Hardware-assisted replication mechanisms & Fast OLAP query response times   &O&X&X& 2011 \\
					&  .Copy-on-write mechanism &  High throughput rates for both OLTP and OLAP &&&& \\
					\hline
					MobiDB \cite{MobiDB:2011}&   Queuing approach & Low latency&O&X&X& 2011\\
					&    & .High throughput rates for both OLTP and OLAP &&&& \\
					&    & .Optimum space overhead &&&& \\
					\hline
					SIE-OBI \cite{Castellanos:2012}&  Data extraction algorithm & Reduce the latency&O&O&X& 2012\\
					& Information correlation information    & Reduce the effort to build data &&&& \\
                    \hline
					
				\end{tabular}}
				\hfill{}
				\caption{A categorization of recent BI systems in terms of modern features.}
				\label{bi_features:tb}
			\end{table*}
		\end{center}
		
			
			While the H-Store system is limited to only OLTP transaction processing, a recent system, called HyPer, can handle mixed workloads from both OLTP and OLAP at extremely high throughput rates using a low-overhead mechanism for creating differential snapshots \cite{Kemper:2011}.
			This system employs the lock-less approach which allows all OLTP transactions to be executed sequentially or on private partitions.
			In parallel to the OLTP processing, the HyPer system executes OLAP queries on the same and consistent snapshot.
			These virtual memory snapshots are created by forking the OLTP process and kept consistent via the implicit operating systems / processor-controller lazy copy-on-write mechanism.
			
			
			Similar to H-Store, MobiDB is a special-purpose main-memory DBMS which guarantees serializability and mixed workloads using the queuing approach \cite{MobiDB:2011}.
			Instead of processing incoming transaction and periodic business queries right away, the MobiDB adds them to a queue and processes them later.
			These requests are first analyzed to estimate how long they would take to be performed.
			Using this analysis and the required guarantees in terms of throughput rate and latency, the MobiDB decides when to execute the queued request adaptively.
			Therefore, the execution times can be estimated quite accurately.
		
		
		To alert business managers of situations that can potentially affect their business, Castellanos et al. \cite{Castellanos:2012} propose a novel platform, called SIE-OBI, that integrates the required functionalities to exploit relevant fast stream information from web .
		The authors proposed novel algorithms which extract and correlate the information obtained from the web with historical data stored in data warehouse to detect the situation patterns.
Only relevant information is extracted from two or more disparate sources of unstructured data, typically one internal slow text stream and one external fast text stream.
		This platform  is created to reduce the time and effort of building data flows that integrate structured and unstructured, slow and fast streams, and analyze them in near-real-time.


\subsection{Data Governance}

\subsubsection{Background}
\begin{figure}
	\centering
	\includegraphics[width=0.38\textwidth]{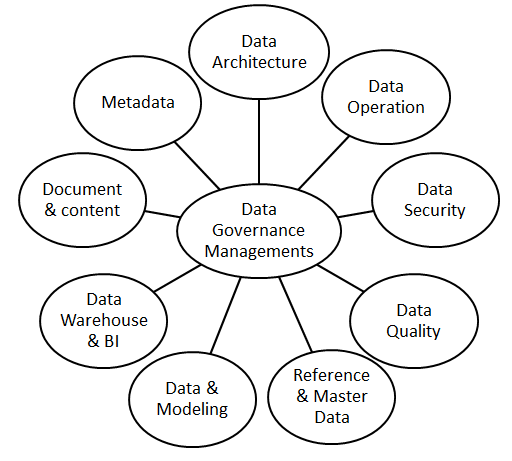}
	\caption{Data governance framework}
	\label{dgframe:fig}
\end{figure}

Data Management International (DAMA I) \cite{DMI2009} defines data governance as ``the exercise of authority and control over the management of data assets, and the planning, supervision and control over data management and use''. Data governance defines roles and responsibilities of the organization to promote desirable behavior in the use of data \cite{WOO2009}. Data governance is differentiated with data management, which involves determining standards for data quality, and making and implementing decisions \cite{KB2010}. It is also differentiated with BI governance, which aims to provide a customized framework for decision-making through the governance of all activities within BI environment \cite{Robert2012}.
DAMA I \cite{DAMA2008} defines ten data management functions as shown in Figure \ref{dgframe:fig}.
Data governance function is high-level planning, supervision, and control over all other functions.

In this section, we focus on only four data management functions related to next generation BI which requires accessing fast to data, utilizing external data, and allowing general users to analyze data.
Data architecture management involves setting data standards, developing and maintaining enterprise data architecture and connecting between the application projects and the architecture.
Data quality management focuses on planning, applying, and controlling activities that apply quality management techniques to measure, assess, improve, and ensure that the data is fit to use.
Data warehousing and business intelligence management focuses on providing decision support data for reporting, query, and analysis.
Meta data management focuses on activities to enable easy access to high quality meta data, such as architecture, integration, control, and delivery.

\subsubsection{Deploying Next Generation BI in Data Governance}
The governance of data is becoming vital to an enterprise as data becomes an asset.
An enterprise derives business value and makes a decision based on the information driven from data.
Thus, the data is needed to be governed to ensure the quality of data which affects directly to the quality of decisions taken by an enterprise \cite{NASCIO1}.

More effective governance of data can result in higher level of decision making.
For achieving effective governance of data, data governance maturity models help an enterprise to understand its data governance and to know what is the anticipated next plan \cite{NASCIO3}.
Several data governance maturity models \cite{PDBA2007} are proposed to guide an enterprise to recognize the level at which its data governance is.
Oracle Corporation \cite{NASCIO3} predicted that a data governance maturity model will assist the enterprise in determining where it is in the evolution of its data governance discipline, identifying the short-term steps necessary to get to the next level, and improving its data governance capabilities \cite{NASCIO3}.
The highest maturity level of Oracle model is integrating data governance with business intelligence.

Next generation BI supports near real-time insights and using the influx of external information which creates a huge amount of data flows and manipulations.
This requires highly matured data governance to provide data quality, integrity, and reliability.
The three properties are crucial for extracting accurate insight via data mining techniques.
For example, in self-service BI, e.g., Tableau and QlikTech, allows users to discover insight from multiple data sources without modelling the data environment and creating complicated ETL processes, which are one of the most difficult and time-consuming tasks in BI.
These new features allow users to access data easily, and get quick results and agile data visualization.
To enable evolution of the next generation BI, data governance is crucial for data reliability of the insight discovered.
For example, in the case of self-service BI, the truth that the end users are able to access and manipulate their own data decreases the reliability of the results of BI \cite{TDWI2015}.
In data governance, useful functions can be considered to ensure reliability such as tracking the data lineage back to the source and creating logs of how the data were manipulated or transformed.
Integrating data governance to the next generation BI, however, has faced some challenges due to the requirements of agile and reliable responses while there are the huge amount of external data and general user participation.

\subsubsection{Data governance challenges}
There are two major features of the next generation BI that affect the data governance model.
In the next generation BI, the decision making should be more effective and expeditious among a large amount of data which are from multiple data sources and formats.
Data from multiple sources, however, make data governance more complicated to control and difficult to manage properly, so that they can also result in ineffective decision making.
If data from different sources conflict, the decision maker has to do much more research and analyze the various data and sources for that data to determine or approximate what is true and accurate, and these processes are costly.
Thus, it is important to manage data across heterogeneous sources and applications in the next generation BI system.

In the next generation BI, especially self-service BI, business users are being involved in decision making procedure.
In general, centralized IT organization and several data stewards have engaged in data governance initiatives and have data management platform metadata repository and a suite of data management tools to handle disparate data.
In advance, they standardize common data definitions for master data and reference data which are broadly shared across many enterprise applications.
When they receive disparate data, they match them to predefined common data definition, determine their quality, define any rules, transform them, and integrate them.
But, in the next generation BI, users also define their own data names, manipulate, or integrate them on their own by using different self-service BI tools.
They might want to upload to the database and share their insight with others.
Business user participation in data process can incur data in chaos since the same data can be transformed and integrated in different ways by a centralized organization and data stewards with data management tools and by business users with self-service BI tools.
Thus, metadata sharing standards are crucial across these participation for common data transformation, common data names, and common integration rules \cite{Mike2014}.

\subsubsection{Data governance model for next generation BI}
The design of the data governance model is classified to centralized v.s. decentralized and hierarchical v.s. cooperative.
The centralized design assigns all decision making authority in a central IT department while the decentralized design distribute the authority over individual business units \cite{WOO2009}.

 \begin{center}
 	\begin{table}[hpt]{
 			\small
 			\begin{tabular} {|p{1.5cm}|p{5cm}|p{1.5cm}|}
 				\hline
 				Roles & Description & Tasks
 				\\ \hline
 				Executive Sponsor & Provides sponsorship, strategic direction, funding, advocacy, and oversight & Consulted
 				\\ \hline
 				Data Quality Board	 & Defines the data governance framework for the whole enterprize and controls its implementation   & Consulted, Informed
 				\\ \hline
 				Chief Steward &  Puts the board’s decisions into practice, enforces the adoption of standards, helps establish DQ metrics and targets & Consulted
 				\\ \hline
 				Business Data Steward & Details corporate-wide DQ standards and policies for his/her area of responsibility from a business perspective & Accountable, Responsible
 				\\ \hline
 				Technical Data Steward & Provides standardized data element definitions and formats, profiles and explains source system details and data flows between systems  & Accountable, Responsible
 				\\ \hline
 				
 			\end{tabular}}
 			\caption{Specification of Data Governance Model for Next Generation BI}
 		\end{table}
 	\end{center}

\section{Next Generation Frontend Architecture}
\label{frontend:sec}
\subsection{Data}
The business insight is obtained from the raw data which is heterogeneous in nature. 
The heterogeneity in the data may be as a result of difference in data sources, the content of data format, type of data or as a result of diversity in data extraction process. 
Depending on the content of study data source can be human generated, machine generated, internal data sources, web and social media, transaction data, biometric data etc.  
Further, the context data format may vary from being structured, unstructured, semi-structured, images, text, videos, audio etc \cite{Thomas2015, Thomas2016b}. 
Considering the data type, there is heterogeneity as meta data, master data, historical and transactional data. 
In the business intelligence study the heterogeneity in data is also contributed by the data extraction method that may depend on the on-demand feeds, continuous feed, real time feeds or time series.

\subsection{BI Analysis}
\par The growing volume of data in many business makes cost effective manual data analysis virtually impossible. 
The use of data mining techniques in business not only handles the volume and variety of data, but also helps to take a proactive knowledge-driven decisions and enhance business intelligence in general. 
Data mining is a broad term that includes a number of process as data modeling techniques, statistical analysis, and machine learning in search for consistent pattern or relationship and determining some predictive information in the data analyzed from large amounts of data \cite{Fayyad:1996}. 
Machine learning leverage to data mining algorithms in order to improve the predictive analysis. 
Machine learning techniques have become popular in BI as they can handle growing volumes and varieties of available data, and makes the  computational processing that is cheaper and more powerful. 
The two broad categories of machine learning algorithms are supervised and unsupervised \cite{Michalski:2013}. 
Supervised learning involves the training data that contain sets of training examples, whereas unsupervised learning draw inference from training data without labeled responses.


\par Business intelligence and analytic shares a mutual relationship. 
Analytics focuses on uncovering data insights that may be beneficial for strategic planning resulting in competitive advantages by analyzing the customer and market behavior in new ways to deliver a quicker real insight. 
The traditional BI analysis process followed a consistent procedure to explore future decisions from the historical data. 
The common tools used to carry out the analysis include sophisticated data analysis tools as OLAP/ROLAP, Machine learning tools and Visualization tools \cite{Khan:2012}. 
These tools provided an access to business users to directly access, interact and visualize the business data without knowing the technical complexity of data retrieval, storage and processing. 
The OLAP is able to provide some analytics functionality as exploring large amounts of data stored in multidimensional database, their relationship, carrying complex computation and visual representation of results from different points of view \cite{Chaudhuri:2011}. 
However, such analysis fails to  give a deeper insight and address question as ``why'' which requires more exploratory perspectives provided by machine learning techniques. 
These techniques are employed to carry out the predictive analytics task, build analytic model at a lower level,  search for predictable behaviors, business rules and look for answers on predicting performance and prescribing specific actions or recommendations. 
The techniques include regression modeling, clustering, neural networks, genetic algorithm, text mining, decision tree, and more.


\subsection{Time series forecasting}
\par In this survey, we will discuss these dynamic decision techniques for time series forecasting. 
The description of time series data includes high dimensionality, volume and continuously evolving. 
The analysis of time series data is a powerful analytics tool as it help to address questions as rate of change in user behavior with time, co-variance between the product, marketing promotion strategies, current trend in product sale, profit monitoring, determine anomalies etc. in nearly all enterprises as sales, manufacturing, mobile companies, hospitals, etc.

\subsubsection{Machine Learning in financial forecasting}
The financial prediction on the stock price is of great interest for the investors as well as the analytics. 
However, the prediction about the current time to buy or sell any stock is not an easy task as the price is influenced by number many factors. 
We list some of the machine learning techniques from the literature commonly used to carry out such predictions.

Support Vector Machine (SVM) is a popular supervised machine learning algorithm, which can be used for classification or regression problems. 
SVM algorithms are able to capture complex relationships between data samples without carrying out difficult transformations. 
Cao et al.\cite{Cao:2003} proposed an application of SVM for financial time series forecasting using single data source. 
The work proposed a SVM with adaptive parameter (ASVM) to handle the structural changes in the financial data. 
The experiment was carried out on five real futures contracts collated from the Chicago Mercantile Market.

The Fuzzy logic methods have been proven effective in dealing with complex  systems containing uncertainties that are otherwise difficult to model. 
In \cite{Chen:2015}  a granular computing is proposed based fuzzy model to improve the accuracy of financial forecast. 
The data source considered for experimentation work considered first-order fuzzy time series model were Taiwan Stock Exchange Capitalization Weighted Stock Index (TAIEX), Dow-Jones Industrial Average (DJIA), S$\&$P 500 and IBOVESPA stock indexes.
 
Hybrid methods are also widely used. A hybrid neurogenetic approach that combined a recurrent neural network  and the genetic algorithm was also used to predict economic growth \cite{Kwon:2007}. 
The recurrent neural network had one hidden layer is used for the prediction model and the genetic algorithm was used to optimize weight.  
The data source used were 36 companies data from  New York Stock Exchange (NYSE) and National Association of Securities Dealers Automated Quotations (NASDAQ). 
Another hybrid approach namely regularized least squares fuzzy support vector regression was proposed to address noise and non-stationarity existing in  financial time series data \cite{Khemchandani:2009}.
Six financial data sets were gathered from Yahoo financial website for IBM, Microsoft, Google Inc., Redhat Software, Citigroups, and Standard $\&$ Poor 500 enterprise. 
The performance  using multi-output support vector regression (MSVR)  considering  interval-valued over short and long horizons \cite{Xiong:2014}. 
The global dataset used for testing were S$\&$P 500 for the US, FTSE 100 for the UK, and Nikkei 225 for Japan.

Today, the stock markets databases are flurried from a wide range of complex data from diverse sources as market data, reference data, exchange, security description, fundamental data as enterprise financial, analyst report, filing etc., and even data from social media that may include blogs, web feeds etc.
Moreover, the financial data is highly dynamic and volatile which raise a need for some integrative approach that combine data from the other sources and contributes towards the accuracy of the forecast.

\subsubsection{Machine learning in sales forecasting}
The prediction of future sale by an enterprise is termed as sales forecasting, which is a part of its critical management strategy. 
The machine learning techniques as Genetic algorithms (GA) are suitable candidates for this task since GAs are most useful in multiclass, high-dimensionality problems where heuristic knowledge is sparse or incomplete.  
Neural networks are also considered as efficient computing models for pattern classification, function approximation and regression problems. 
The sales forecasting problem for printed circuit board (PCB) sales addressed in [\cite{Chang:2005}], where the model is built by integrating GAs and Neural Network. 
The study was carried out for PCB electronic industries in Taiwan, where the feature of data included monthly sales amount, total production square measures, etc.  
The PCB sale is studied [\cite{Hadavandi:2011}], using integrated genetic fuzzy systems (GFS) and data clustering. 
The approach was experimented on the PCB data source with parameters as pre-processed historical data, Consumer price index, Liquid crystal element demand and Total production value of PCB.
Kuo et al. \cite{Kuo:1999} proposed a hybrid machine learning system with fuzzy neural network on locally chain supermarket data. They method enabled incorporating expert knowledge in the forecasting. 
In one example of automobile sale forecast \cite{Wang:2011}, adaptive network-based fuzzy inference system was considered, which included several economic variables as current automobile sales quantity, coincident indicator, leading indicator, wholesale price index and income for prediction.
 The data analyzed was whole automobile market in Taiwan that included sales data corresponding to sedan, small commercial vehicle, and large commercial vehicle.
The sales forecasting is very complicated owing to influence by internal and external environments.

\subsubsection{Machine learning in heathcare}
The time series data are studied by the healthcare enterprises using machine learning to understand existing patterns that help the administration to make strategic decisions. 
Some prediction based on the time series data includes outpatient visits, and customer behavior in choosing hospital
Chang et al.\cite{Chang:2008} propose a fuzzy logic based approach based on weighted transitional matrix to forecast the outpatient (patient who receives medical treatment without being admitted to a hospital) visits. 
The forecasting is important as effective prediction helps the administration to manage operation, distribute resources and other aspects. 
The build model was tested on the data gathered from the department of internal medicine in a hospital. 
The data had two features to be monitored, month of the year and the number of outpatient. 
Another similar study was carried out by Hadavandi et al. \cite{Hadavandi:2012} were a hybrid model was built that combine genetic algorithm with fuzzy rule based learning to forecast outpatient visits. 
The data were collected from the department of internal medicine in a hospital in Taiwan and four big hospitals in Iran. 
The data features being the month and the number of outpatient. 
In a different application, neural network techniques were used to make predictions about the consumers behavior in choosing hospital \cite{Lee:2008}. 
The results are useful as the hospital operating environment is getting more competitive. 
The data feature considered were  cost of medical care, accessibility, parking, hospital reputation, doctor reputation, doctors medical skill, modern equipment, etc

\subsection{Improving BI via integrative data analysis}

In today’s enterprise, data are created from multiple source. 
The data from multiple sources can provide an insight to increase productivity, improve policy making, support performance measurement,  and can help in strategic planning.
These insights can help achieve benefits as improved customer satisfaction, quality improvement,  increased accessibility and analysis of information, timeliness and better information utilization.  
The need to handle the heterogeneous data and automated analytics algorithm can be resolved by doing predictive analysis. 
It supports the study of the  integrated data  using machine learning  algorithm that continuously evolve the accuracy of predictive models and enable it to adjust.

An integrative analysis task involves machine learning techniques for the integration of the available training data from different data sources in order to better analysis and generate a proactive response. 
A single data source may not contain all required information about the data object. 
Thus, when we combine multiple sources of information for a particular data object it helps to add on some different or missing information that may lead to a better prediction accuracy. 
Integrating data from multiple sources and making decisions from these combined sources is becoming common to enhance the prediction performance for different applications as in bio-informatics, image classification, stock market, etc. The two most common integrative analysis are Multiple Kernel Learning and the Bayesian Network (BN)\cite{Thomas2015}. 
Tensor decomposition based data mining algorithms is also promising. Tensor are high dimensional array which naturally allows multiple source of the same component can be integrative represented and analyzed in both time series data as well as static data \cite{Sael2015,Jeon2016a,Shin2017}. However, we will mainly focus on applications of MKL. 

\subsubsection{Multiple kernel learning}
Multiple kernel learning (MKL) refers to a set of machine learning methods that use a predefined set of kernels, where kernel selection depend on the notions of similarity that may exist in the data source.
MKL learn an optimal linear or non-linear combination of kernels as part of the algorithm that result in data integration. 
MKL algorithms have been developed for supervised, semi-supervised, as well as unsupervised learning.

\par In  stock price prediction it is observed that relying on the study of the time series historical data is not sufficient, rather by considering multiple sources of information such as news and trading volume one can significantly improve the stock market volatility predication \cite{Wang:2012}. The experiments carried out on HKEx 2001 stock market datasets shows that the use of multiple kernel learning approach on the multiple data sources results in higher accuracy and lower degree of false prediction as compared to single source data.


\par The work \cite{DENG:2011}, shows that analyzing communication dynamics on the internet and using stock price movements may provide some new insights into
relations between stock prices and communication patterns. They use MKL to combine information from time series data, stock price and stock volume with other data source: news and comments that included the frequency of News, frequency of the comments, average Length/ Standard Deviation of length of comments, number of Early/Late response etc. The experiment was tested for the stock of Amazon, Microsoft and Google and it was found that MKL prediction model outperforms other
baseline methods Mean Absolute Error (MAE), Mean Absolute Percentage Error (MAPE) and Root Mean Square Error(RMSE).

\par Integrative data approaches are applied to draw inferences from the biomedical data.  In \cite{Yu:2010}, the paper reports the advantage of the MKL integrative approach that thoroughly combining complementary information from the heterogeneous data sources over the sparse integration method for the biomedical data.  The experiments were carried out for different applications as  disease relevant gene prioritization by genomic data fusion, Prioritization of recently discovered prostate cancer genes by genomic data fusion, Clinical decision support by integrating microarray and proteomics data, Clinical decision support by integrating multiple kernels integration of genome-wide data for clinical decision support in cancer diagnosis, and Computational complexity and numerical experiments on large scale problems. An application is also seen in \cite{Speicher:2015}, where the different biological measurements are integrated using the regularized unsupervised multiple kernel learning to find cancer subtypes. In \cite{Savage:2010}, bayesian networks are considered for data integration of two different biological data (gene expression and transcription factor binding (ChIP-chip)) for discovering the transcriptional modules.


\par Integrative data analysis finds applicability in high-dimensional hyperspectral classification. In satellite remote sensor application \cite{Tuia:2010}, MKL approach proves beneficial because of the high-dimensional feature induced by using many heterogeneous information sources. The approach is based
on the automatic optimization of a linear combination of kernels
dedicated to different meaningful sets of features as groups of bands, contextual or textural features, or bands acquired by different sensors. The results obtained showed a good performance of the method in image classification, when multispectral, hyperspectral, contextual or multi-sensor information were used.
 The urban classification using MKL was carried \cite{Gu:2015},  to integrate heterogeneous features from two data sources, i.e., spectral images and LiDAR data. The features from spectral images are good at identifying ground truth such as trees, grass, and soil; whereas features from LiDAR data perform better for classes with flat surfaces such as buildings. The classification model build by combining the data source with complementary relationship significantly improve the classification accuracy.

\section{Use Cases}
\label{usecases:sec}
\subsection{Use case: BI in Healthcare}
The wealth of available data in healthcare will continue to increase. This together with raises the demand for improved quality of patient care, necessitates the improvement of Healthcare BI.  
This rising demand can be best addressed by considering the business intelligence technological solutions of data acquisition, storage, interpretation and evaluation.

\subsubsection{Heterogeneous data in healthcare}
Healthcare deals with huge heterogeneity in the data coming from different sources. The  major challenges involved in managing healthcare data are format, structure and complex nature. 
Health care data occurs in different formats as numeric, textual, digital, images, videos, multimedia etc. Electronic health record (EHR) holds hundreds of rows of textual and numerical data corresponding to patient demographics, progress notes, vital signs, medical histories, diagnoses, radiology images, medications, lab and test results. 
In healthcare, the data is structured and unstructured. 
Structured data refer to the lab or patient demographic data that is consistent and stored in a pre-defined format. 
The unstructured data are nonuniform and can be of great value in analyzing the patient data. 
These include clinical notes, audio voice dictations, email messages and attachments, text message, online video and typed transcriptions.
The presence of structured and unstructured data makes the healthcare data complex  to process and increase further as the number of variable increases.

\subsubsection{Use of sensor in healthcare} 
Business intelligence in healthcare provides a wide range of analytics to improve the decision making process related to both patient and performance that covers many functional areas, including resource planing, care delivery, patient accounting, financial and revenue cycle.
The technological advancement has allowed a greater accessibility to data. 
The use of sensors in healthcare produces large volumes of data continuously over time. 
An application can be seen in the Intensive Care Unit (ICU) were sensors are used to monitor the current state of patients, it includes ECG, EEG, blood pressure monitors, respiratory monitors.

\subsubsection{ICU case study}
Intensive care units has a data rich environment with multiple source of continuous data originate from medical devices which includes electroencephalogram, Bedside Monitors, Brain Tissue Oxygen Monitor,  Central Venous Catheters (CVC), Clinical Information Systems and ventilators,  resulting in several kilobits of data each second per patient. 
A patient in severe health state are often monitored by number of body sensors connected to  monitoring devices producing large volumes of physiological data, along with comprehensive and detailed clinical data and minute-by-minute trends for the patient. 
Another, ICU patient monitoring involves the system to generate some alerts or alarms when the physiological state of the patient shows the detection of relevant abnormalities or changes in a patient's condition. 
The physicians use certain thresholds, which when exceeds triggers the in alarm.  
However, such simple alerting schemes may result in large number of false alarms, example,Tsien and Fackler \cite{Tsien:1997} found that 92$\%$ of alarms were false alarms in their observation in a pediatric intensive care unit,  in \cite{Siebig:2010}, authors digitally recorded all the alarms for 38 patients on a 12-bed medical ICU and retrospectively assessed their relevance and validity: Only 17$\%$ of the alarms were relevant, with 44$\%$ being technically false.

In case of medical emergency, it is critical for the patient to receive the medical aid at the earliest. 
The recent advancement in the sensor technology helps to achieve this goal by monitoring patient crucial parameters such as heart beat, body temperature, blood monitoring, EEG etc.  
In the given scenario, business intelligence can play a key role by carrying out the predictive analysis  by integrating the sensor data with the available traditional data.   
The processed data that give the vital statistics of the patient can be transmitted to the doctor to guide the paramedic traveling with the patient. 
In addition, this information can be used to take appropriate action as soon as the patient arrives at the hospital.

The existing complexity in the data makes the existing approaches may not be suitable to manage data in healthcare. 
In healthcare business rules and definitions are volatile and may change  over a period of time. 
This calls for a solution that is based on agile  approach and can handle data from multiple sources, the different format, the structured and unstructured data and manages the complexity within an ever-changing regulatory environment.
Thus, an intelligent machine learning algorithms are required to find a coherent meaning from disparate data to process heterogeneous data captured through different sensors.

\section{Conclusion}
\label{conclusion:sec}
We have reviewed traditional and next generation business intelligence and analytic in a holistic view. 
The three-tier traditional BI architecture is still valid. 
However, it is not sufficient in providing real-time analysis, situation aware, and self-service capabilities. 
The next generation BIs, i.e., operational BI, situation BI, and self-service BI, are each focusing in realization of the three capabilities that becoming more important in BI as the data becomes business assets. 
We looked at challenges and enabling technologies for the three type of BIs in the back-end perspective. We also pointed out data governance is critical in the next generation BIs. 
In the front-end, we looked at data analytic methods focusing on time series forecasting and integrative data analysis. 
We also looked at healthcare uses case and showed that next generation BI and analytics can save lives. 
Next generation BI\&A is at its beginning and there remains several challenges in both the back-end architecture and in the front-end analysis.

\bibliographystyle{ieeetran}
\bibliography{bibliography}


%




\end{document}